\documentclass{IEEEcsmag}

\usepackage[colorlinks,urlcolor=blue,linkcolor=blue,citecolor=blue]{hyperref}
\expandafter\def\expandafter\UrlBreaks\expandafter{\UrlBreaks\do\/\do\*\do\-\do\~\do\'\do\"\do\-}
\usepackage{upmath,color}

\jvol{XX}
\jnum{XX}
\paper{8}
\jmonth{May 2025}
\jname{METACOG-25: 2nd Workshop on Metacognitive Prediction of AI Behavior}
\jtitle{METACOG-25: 2nd Workshop on Metacognitive Prediction of AI Behavior}
\pubyear{}

\begin{document}

\sptitle{\phantom{.}}

\title{Metacognition in Content-Centric Computational Cognitive C$^4$ Modeling}

\author{Sergei Nirenburg, Marjorie McShane, Sanjay Oruganti}
\author{}\affil{The Language-Endowed Intelligent Agents Lab, Rensselaer Polytechnic Institute, Troy, NY, 12180, USA}

\markboth{METACOG-25}{METACOG-25}

\begin{abstract}\looseness-1   For AI agents to emulate human behavior, they must be able to perceive, meaningfully interpret, store, and use large amounts of information about the world, themselves, and other agents. Metacognition is a necessary component of all of these processes. In this paper, we briefly a) introduce content-centric computational cognitive (C$^4$) modeling for next-generation AI agents;  b) review the long history of developing C$^4$ agents at RPI’s LEIA (Language-Endowed Intelligent Agents) Lab; c) discuss our current work on extending LEIAs’ cognitive capabilities to cognitive robotic applications developed using a neuro symbolic processing model; and d) sketch plans for future developments in this paradigm that aim to overcome underappreciated limitations of currently popular, LLM-driven methods in AI.
 
\end{abstract}

\maketitle
Metacognitive abilities are a key prerequisite for making AI agents full-fledged members of human-AI teams. AI agents must use metacognition both for \textit{introspection} and for \textit{mindreading}—that is, understanding the knowledge, reasoning, intentions, skills, personality traits, and preferences of themselves and their teammates. Core prerequisites for introspection and mindreading are maintaining and dynamically enhancing a) the agent’s ontological model of the world and the agents in it; b) resources that link elements of perception with the agent’s mental models (e.g., a lexicon that links words and phrases to ontological concepts); and c) the agent’s memories of past experiences of perception interpretation, reasoning, and action. All this content supplies essential metacognitive heuristics for the agent’s decisions.

It cannot be overemphasized that semantically interpretable knowledge resources are essential to an agent’s ability to select appropriate actions and explaining why those actions were chosen. This is the \textit{content-centric} aspect of C$^4$ modeling. Moreover, interpreted knowledge facilitates an agent’s instructing and being instructed by other agents through show and tell, the way people are taught in everyday situations and in all manner of training environments.

Crucially, interpreted knowledge resources support \textit{a variety} of computational approaches to realizing metacognitively endowed AI agents – rule-based and machine learning-based ones as well as hybrid, so-called neurosymbolic approaches. 

\section{A Brief Survey of Metacognition in Agents Developed Using C$^4$ Modeling}
\vspace{5pt}
\chapteri{W}ei et al. \cite{wei2024metacognitive} characterize metacognition as supporting the following four capabilities (the definitions are ours): \textbf{transparency,} which involves an agent’s explaining its reasoning and decision-making; \textbf{adaptability} to novel situations and in support of lifelong learning; \textbf{reasoning,} including its self-aware aspects; and \textbf{perception,} which requires interpreting the output of perception-oriented technologies. This taxonomy is incomplete, especially if we take into account cognitive robotics. For this, a fifth capability must be added: \textbf{action,} both physical and verbal, which agents must carry out within their teams.
In the LEIA lab, we have been developing all of the abovementioned capabilities within the C$^4$ modeling framework. Sample prototype applications are the Maryland Virtual Patient (MVP) system for training medical students \cite{mcshane2021linguistics}(Ch. 8); a virtual vehicle agent \cite{mcshane2024agents}(Sec. 7.1.5); and several simulated human-robot team applications based on the HARMONIC cognitive-robotic architecture \cite{oruganti2024harmonic}.
\vspace{-12pt}
\subsection{Transparency}
In all our systems, (a) the output of all system modules is available, in human-legible form, for inspection, and (b) a special module is devoted to generating explanations, in plain English, of the reasons for agent decisions.
\vspace{-12pt}
\subsection{Adaptability}
When virtual patients in MVP engage in dialog with human users, they mindread them, taking into account their personality traits, physical and mental states, and levels of domain knowledge. In all of our systems, agents engage in lifelong learning of new ontological concepts and new lexical material through dialog with teammates. This is made possible by our extensive work on deep natural language understanding that uses stored knowledge resources for bootstrapping (see, e.g., \cite{mcshane2024agents} Ch. 7).
\vspace{-12pt}
\subsection{Reasoning}
LEIAs engage in reasoning when interpreting input, deciding on instantiating and prioritizing goals, selecting plans, carrying out plans, dealing with disturbances, and choosing how to implement individual actions within the plans. All these tasks involve heuristic decision functions whose argument sets include values of a number of metacognitively-related metaparameters. For example, if the computational cost of determining a parameter value in a decision function is too high, then the function can be run without that feature, albeit with a lower \textit{confidence} in the resulting decision. Confidence is, in turn, computed using metaparameters including \textit{vagueness} and \textit{incompleteness} of sensory input. Confidence is among the determinants of \textit{actionability} – that is, whether the agent believes it is licensed to act on an incomplete understanding of input, given an application’s requirements (for detailed discussions of actionability, see \cite{mcshane2021linguistics,mcshane2024agents,oruganti2024harmonic}).
\vspace{-12pt}
\subsection{Perception}
In all our systems, the results of perception are interpreted in terms of the system’s knowledge resources. Interpretation routinely takes into account metacognitive aspects, such as the agent’s history, its mindreading of other agents, etc. A good example of the use of metacognition in perception is the LEIAs’ ability to recover from ill-formed language utterances and detect cognitive biases in others (see relevant discussions throughout \cite{mcshane2021linguistics}, especially Ch. 3 and Ch. 4, and Sec. 8.2).
\vspace{-12pt}
\subsection{Action}
In all our systems, LEIAs generate verbal actions to communicate with teammates. LEIAs not only produce an English rendering of the underlying thought but also select a style and word choice that reflects mindreading of teammates’ beliefs, intentions, emotions, and personality traits as well as their shared history. Thus, when a virtual patient in MVP comes to a repeat visit to a particular doctor and the doctor asks, “How are you?” the LEIA judges it appropriate to respond with the comparative “I’m feeling better.”

\section{Evolution of the Computational Infrastructure for C$^4$ modeling}
Originally we implemented LEIAs as predominantly rule-based systems. But in light of the technological leap offered by LLMs, we recently switched to a hybrid, neurosymbolic infrastructure. However, our approach to hybridization differs from most current integration proposals (see \cite{besold2021neural} for a survey), which focus on LLMs and use limited knowledge-based support in an effort to boost performance. Our approach is the opposite: We focus on C$^4$ modeling with the goal of producing trustworthy agents and integrating LLMs as a means of improving system performance. To date, we have incorporated LLMs in two components of LEIAs – language generation and life-long learning through understanding.

Unlike LLMs, C$^4$ agents generate text intentionally as a step in consciously pursuing a goal. This process involves both the selection of the content to be conveyed and the choice of how to actually say it in English. We use knowledge-based methods to select the content and generate multiple candidate sentences to convey it. Then we use an LLM to decide which of those sentences is best in the context. This is precisely the kind of task that LLMs are good for because it requires a mastery of how the surface level of language works without the need to take responsibility for its content (see \cite{mcshane2024agents}, Sec. 4.3). We have tested a variation on the above capability in a system for automatic authorship anonymization \cite{mcshane2025neurosymbolic} in which LLMs helped to filter out atypical textual formulations and offered additional text paraphrase solutions when the knowledge-based engine failed to adequately anonymize a text.

To implement lifelong learning through understanding, LEIAs use their available resources and processors to learn new, and improve existing, lexicon entries and ontological concepts by understanding natural language texts or inputs from human or robotic instructors. Our team’s early implementations of this process \cite{nirenburg2007learning, mcshane2015learning, nirenburg2017toward, nirenburg2018toward, nirenburg2021overcoming} were rule-based. The approach we are now working on incorporates LLMs to enhance the efficiency of the learning process by filtering the lexical material for LEIAs to interpret during the learning process. The algorithm for this process, described in detail in \cite{mcshane2024agents} (Ch. 7), is currently being implemented in an application of the HARMONIC cognitive-robotic architecture \cite{oruganti2024harmonic}.\footnote{We plan to demonstrate this capability in a demo at the conference.}

\section{Conclusion}
This paper argues that content-centric computational cognitive (C$^4$) modeling is the most promising methodology for building trustworthy AI agents that are self-aware and capable of human-level explanations. Only such agents are fit for truly critical applications in defense, health, finance, etc. Metacognition is an integral feature of C$^4$ modeling, as illustrated by the above examples of  C$^4$-based systems the RPI LEIA lab has built. We have also demonstrated that C$^4$ modeling can be implemented in a variety of computational infrastructures, including the novel neurosymbolic one we are implementing. In the immediate future we intend to demonstrate that our approach to lifelong learning through understanding will remove the so-called “knowledge bottleneck” and will facilitate the development of flexible and reliable agents and robots that can become full-fledged members of human-AI teams.

\def\refname{REFERENCES}
\bibliographystyle{IEEEtran}

\vspace{-8pt}
\begin{IEEEbiography}{Sergei Nirenburg}{\,} is a Professor of Cognitive Science and Computer Science at RPI, and Co-Director, of RPI’s Language-Endowed Intelligent Agents (LEIA) Lab. He has worked in the areas of cognitive science, artificial intelligence, and natural language processing for over 45 years and led many R\&D teams. His lab develops explanatory theories of human cognitive functioning and implements them in artificial intelligent agents operating in human-agent teams. Dr. Nirenburg has co-authored 4 and edited 6 books and published over 250 peer-reviewed scholarly articles. He chaired program committees at over a dozen NLP- and MT-related conferences and served as a director of two NATO-sponsored Advanced Studies Institutes. You can contact him at nirens@rpi.edu
\end{IEEEbiography}
\vspace{-8pt}
\begin{IEEEbiography}{Marjorie McShane}{\,} 
 is a Professor of Cognitive Science and Co-Director of the Language-Endowed Intelligent Agents (LEIA) Lab at RPI. Her research focuses on computational cognitive modeling of intelligent agents, with emphasis on knowledge, learning, and communication. She recently coauthored two influential books with MIT Press: Linguistics for the Age of AI (2021) and Agents in the Long Game of AI: Computational cognitive modeling for trustworthy, hybrid AI (2024). Her extensive publication record includes two single-authored books, three edited volumes, 46 journal articles and book chapters, and over 60 peer-reviewed conference papers. Contact her at mcsham2@rpi.edu.
\end{IEEEbiography}
\vspace{-8pt}
\begin{IEEEbiography}{Sanjay Oruganti} {\,}is a Research Scientist at the Language Endowed Intelligent Agents (LEIA) lab at RPI. His research interests include cognitive robotics, emergent behaviors in multi-robot systems, and knowledge-sharing frameworks for robotic teams. He received his PhD in Electrical and Computer Engineering from the University of Georgia, USA. Dr. Oruganti is a Senior Member of the IEEE. Contact him at orugas2@rpi.edu.
\end{IEEEbiography}

\end{document}